\documentclass{bmvc2k}

\usepackage{graphicx}
\usepackage{subcaption}
 \usepackage{booktabs}
 \usepackage{multirow}
 \usepackage{wrapfig}
 \usepackage{float}
 \usepackage{enumitem} 
\newlist{flatitem}{itemize}{1}
\setlist[flatitem]{label={}, leftmargin=0pt, itemsep=0pt, topsep=0pt}

\definecolor{mySectionColor}{rgb}{0, 0, 0} % black
\makeatletter
\renewcommand\section{\@startsection {section}{1}{\z@}%
  {-3.5ex \@plus -1ex \@minus -.2ex}%
  {2.3ex \@plus.2ex}%
  {\normalfont\Large\bfseries\flushleft\textcolor{mySectionColor}}}

\renewcommand\subsection{\@startsection{subsection}{2}{\z@}%
  {-3.25ex\@plus -1ex \@minus -.2ex}%
  {1.5ex \@plus .2ex}%
  {\normalfont\large\bfseries\flushleft\textcolor{mySectionColor}}}
\makeatother

\title{SegDebias: Test-Time Bias Mitigation for ViT-Based CLIP via Segmentation}

\begin{document}

\begin{center}
    {\LARGE \bfseries SegDebias: Test-Time Bias Mitigation for ViT-Based CLIP via Segmentation \par}
    \vspace{1em}
    {\normalsize \textbf{Fangyu Wu} \quad \textbf{Yujun Cai} \par}
    \vspace{1em}
    \hrule
    \vspace{1em}
\end{center}

\begin{abstract}
Vision language models such as CLIP have shown remarkable performance in zero shot classification, but remain susceptible to spurious correlations, where irrelevant visual features influence predictions. Existing debiasing methods often require access to training data and explicit group labels to perform fine-tuning or adjust embeddings, which limits their practicality in real-world settings. Test-time methods attempt to avoid this constraint, but many still depend on prior knowledge of dataset specific biases, limiting their generalizability in open set settings. In this work, we propose a test-time debiasing method for ViT based CLIP models that requires no additional training or assumptions of bias annotations. Our approach uses a pretrained segmentation model to isolate the target visual attribute, then adjusts the non target regions so that their embeddings are uniformly similar to all class specific text prompts. This procedure removes unintended bias signals from confounding visual regions while preserving the target attribute. Experiments on Waterbirds and CelebA show that our method outperforms existing test-time debiasing approaches in both group robustness metrics and Attention IoU. These results demonstrate the effectiveness of segmentation guided interventions for scalable and annotation free bias mitigation in vision language models.
\end{abstract}

%-------------------------------------------------------------------------
\section{Introduction}
\label{sec:intro}
Recent advances in vision-language models (VLMs), such as CLIP ~\cite{Radford21}, have enabled strong zero-shot classification performance by aligning large-scale image and text pairs. However, growing evidence shows that these models are vulnerable to spurious correlations, where unintended visual patterns influence predictions and lead to biased or unreliable behavior in real-world settings~\cite{Luo24, Bansal23, Kim23}. When classifying waterbirds versus landbirds, for example, CLIP often focuses on background elements (water or land) instead of the birds themselves. This misplaced attention leads to poor performance in minority groups that break these common associations, creating significant performance gaps between demographic subgroups in datasets such as CelebA~\cite{CelebA}, where attributes such as hair color become inappropriately linked to gender-specific facial features.

Existing solutions to this problem face substantial practical limitations. Training-based approaches such as FairCLIP~\cite{Luo24} and Contrastive Adapters~\cite{zhang2022contrastive} require group labels and training data, resources frequently unavailable in real-world settings. While test-time methods like DebiasedPrompt~\cite{Chuang23} and BendVLM~\cite{Gerych24} avoid retraining costs, they still rely on explicit knowledge of bias-inducing attributes, severely limiting their usefulness in open-set environments where biases may be unknown or complex.

\begin{figure*}[h]
  \begin{center}
    \begin{tabular}{cccc}
      \includegraphics[width=3cm]{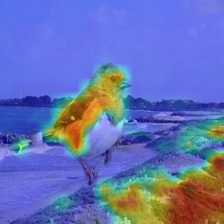} &
      \includegraphics[width=3cm]{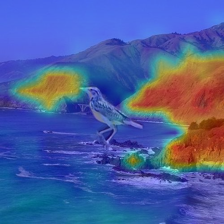} &
      \includegraphics[width=3cm]{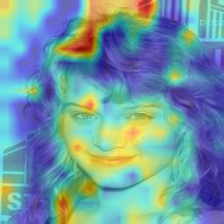} &
      \includegraphics[width=3cm]{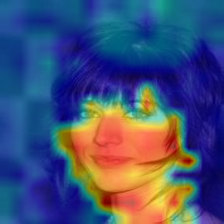} \\
      (a) Mixed focus &
      (b) Erroneous focus &
      (c) Mixed focus &
      (d) Erroneous focus
    \end{tabular}
  \end{center}
  \caption{Attention-weight feature maps (Grad-CAM) from a ViT-based CLIP model on two binary classification tasks: waterbird vs.\ landbird (panels a–b) and blond-hair vs.\ dark-hair (panels c–d). Panels (a) and (c) show mixed attribution where both target and non-target regions are emphasized; panels (b) and (d) illustrate failure modes where attention is misdirected exclusively to irrelevant features. (highlighted colors from red to green indicate increasing levels of attention map values, while blue areas represent minimal or no focus).}
  \label{fig:gradcam-qualitative}
\end{figure*}

The core challenge lies in identifying and addressing these spurious correlations without prior knowledge of their nature. Through careful analysis of CLIP's attention patterns, we discovered that these models frequently highlight image regions irrelevant to the classification task, focusing on water backgrounds when identifying birds, or on facial features when classifying hair color. This attention misalignment directly causes the performance disparities between groups shown in Figure \ref{fig:gradcam-qualitative}.

Based on these insights, we developed SegDebias, a straightforward yet effective test-time approach that addresses the root cause of bias by directly intervening in the visual attention space. Our method operates without any knowledge of dataset-specific biases or protected attributes, making it widely applicable across domains. Instead of trying to identify all possible spurious correlations, SegDebias uses semantic segmentation to isolate the target attribute (such as a bird in waterbird classification), then carefully adjusts the non-target regions to have equal similarity to all candidate text embeddings. This neutralization effectively removes their ability to bias the model's prediction while preserving the important visual information from the target attribute.

%We demonstrate the effectiveness of our method by benchmarking against state-of-the-art test-time debiasing techniques. Our approach not only improves \textbf{Attention-IoU} alignment between model attention and the target visual region, but also achieves the highest \textbf{worst-group accuracy} across all evaluated test-time methods. These results highlight the robustness of our segmentation-guided intervention. To summarize, the primary contributions of our work are:
Extensive experiments on Waterbirds~\cite{Sagawa20Waterbirds} and CelebA~\cite{CelebA} demonstrate that SegDebias substantially outperforms existing test-time debiasing methods, achieving the best worst-group accuracy while significantly reducing performance gaps between groups. By measuring Attention-IoU~\cite{Serianni25}, we show that our approach successfully redirects CLIP's attention toward semantically meaningful regions, providing a clear explanation for its improved fairness.
To summarize, the primary contributions of our work are:
\begin{enumerate}[itemsep=0.2pt, topsep=2pt]
  \item We introduce a segmentation-guided test-time debiasing method for ViT-based CLIP that requires no prior knowledge of dataset biases or protected attributes.
  \item We use the Attention-IoU metric to reveal limitations in CLIP’s attention patterns and show how our method improves semantic alignment.
  \item We demonstrate that our approach outperforms existing test-time debiasing methods across standard benchmarks in group robustness.
\end{enumerate}

\section{Related Work}
{\bf Representation-Level Debiasing Techniques}. Recent efforts to mitigate bias in vision-language models (VLMs) like CLIP~\cite{Radford21} have predominantly focused on modifying the model's output embeddings, either through fine-tuning or test-time adjustments. FairCLIP~\cite{Luo24} introduces an optimal-transport-based approach, reducing the Sinkhorn distance between protected-group distributions to balance performance and fairness, but depends on labeled data and extra training. Likewise, FairerCLIP~\cite{Dehdashtian24} projects CLIP’s image and text embeddings into a reproducing kernel Hilbert space (RKHS) and solves a closed-form optimization to remove bias. Contrastive Adapters~\cite{zhang2022contrastive} train lightweight MLP-based adapters inserted into both the image and text towers of CLIP to adjust their embeddings. In contrast, test-time only methods seek to correct biases at inference: DebiasedPrompt~\cite{Chuang23} adjusts CLIP’s input prompts to cancel known spurious directions, but requires explicit knowledge of dataset biases whereas Bend-VLM~\cite{Gerych24} adjusts candidate text embeddings by performing two-stage, per-input nonlinear debiasing: orthogonalizing to local attribute subspaces and equalizing similarity across protected groups. Although these test-time approaches advance CLIP’s fairness, they assume prior access to target datasets with known protected attributes, making them ill-suited to open-set, test-time scenarios where confounders are unknown or complex.

\noindent
{\bf Attention-IoU: Bias Detection through Attention Maps} In addition to these debiasing strategies, Attention-IoU~\cite{Serianni25} introduces an anatomy of bias inside the CLIP decision process by comparing model attention maps to ground truth masks or other attribute maps. Using Grad-CAM, the authors define real-valued Intersection-over-Union between attention and mask or between two attention maps, yielding the mask score and heatmap score respectively. In Waterbirds~\cite{Sagawa20Waterbirds}, the mask score accurately tracks background-induced bias, while in CelebA~\cite{CelebA}, the framework reveals that classifiers often rely on gendered facial features (e.g., eyes, nose) when predicting attributes like Wearing Lipstick or Blond Hair, even when those features are not directly relevant, highlighting spurious correlations beyond what label statistics alone can reveal. These findings reveal that even without output errors, CLIP’s internal focus can betray biased shortcuts, suggesting that segmentation-guided interventions at test-time could potentially steer attention back to semantically relevant regions.

\section{Method}
\subsection{Motivation}
As discussed in Section \ref{sec:intro}, Figure~\ref{fig:gradcam-qualitative} shows that the ViT encoder in CLIP~\cite{Dosovitskiy21ViT} often over-weights spurious, non-target regions such as background textures or contextual cues when computing visual embeddings. In some cases, even when the target object (e.g., a bird) is masked out, the background alone can yield high cosine similarity to an incorrect class prompt. This behavior suggests that confounding factors, although not explicitly modeled, can disproportionately influence CLIP’s zero-shot predictions.

To quantitatively analyze the influence of non-target visual regions on CLIP's zero-shot predictions, we conducted analysis on two benchmark datasets: Waterbird~\cite{Sagawa20Waterbirds} and CelebA~\cite{CelebA} (see Section \ref{sec:4.1} for more details). In Waterbirds, the task is to classify images as “waterbird” or “landbird”; in CelebA, the goal is to classify hair color as either “blond” or “dark.” For each dataset, we segmented images into two regions: the target attribute (bird or hair) and the non-target region (background or face). We then passed the non-target regions through CLIP’s visual encoder to obtain embeddings.

We computed the cosine similarity difference between these non-target embeddings and the class text prompts as:
{\setlength{\abovedisplayskip}{4pt}
 \setlength{\belowdisplayskip}{4pt}
\begin{equation}
\Delta_{\text{non-target}} = \text{sim}(E_{\text{non-target}}, E_{\text{prompt}_1}) - \text{sim}(E_{\text{non-target}}, E_{\text{prompt}_2})
\end{equation}
}
and similarly for the full image embeddings:
{\setlength{\abovedisplayskip}{4pt}
 \setlength{\belowdisplayskip}{4pt}
\begin{equation}
\Delta_{\text{full}} = \text{sim}(E_{\text{full-image}}, E_{\text{prompt}_1}) - \text{sim}(E_{\text{full-image}}, E_{\text{prompt}_2})
\end{equation}
}

\begin{figure}[h]
  \begin{center}
    \begin{tabular}{cccc}
      \includegraphics[width=6cm]{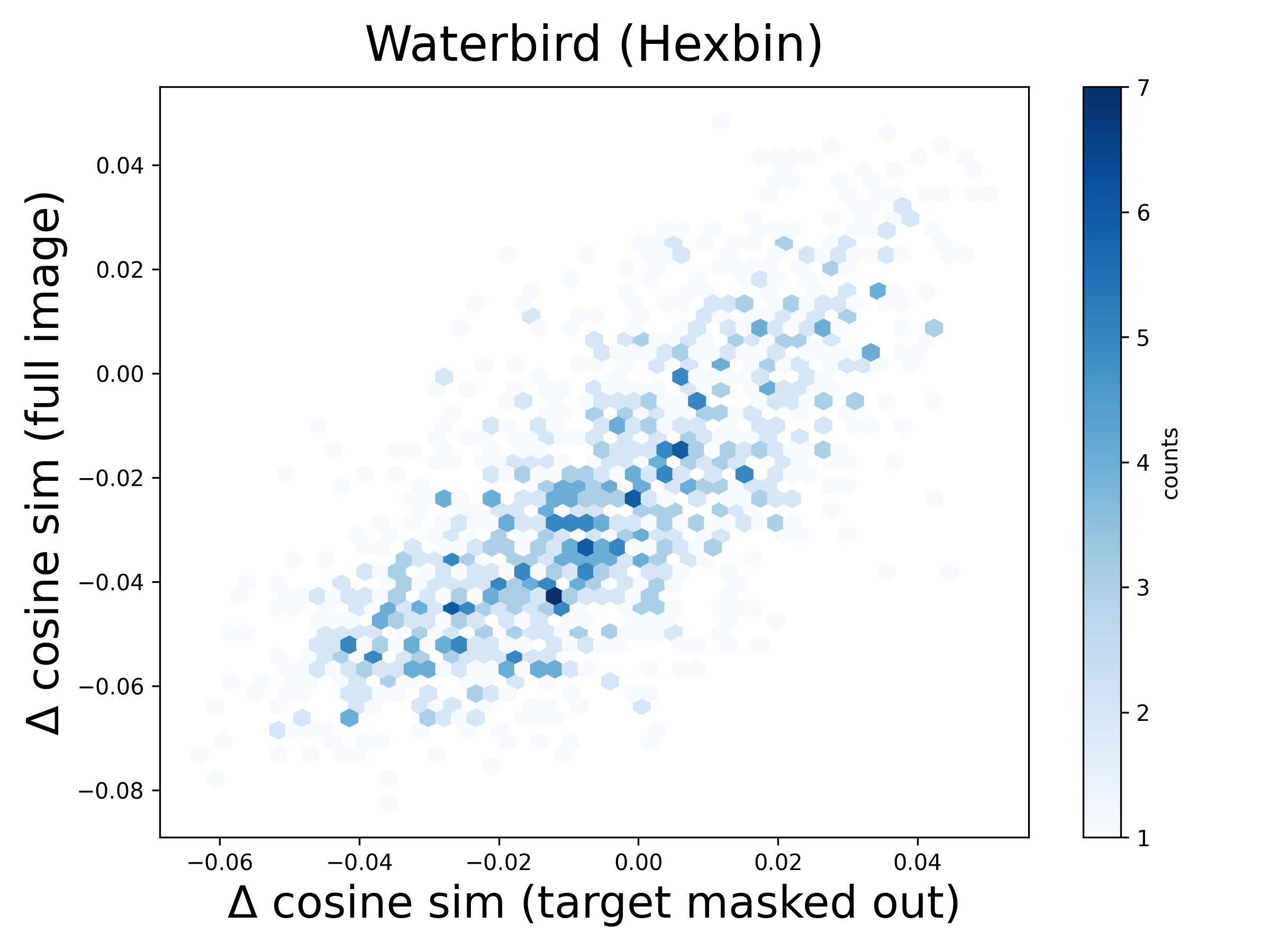} &
      \includegraphics[width=6cm]{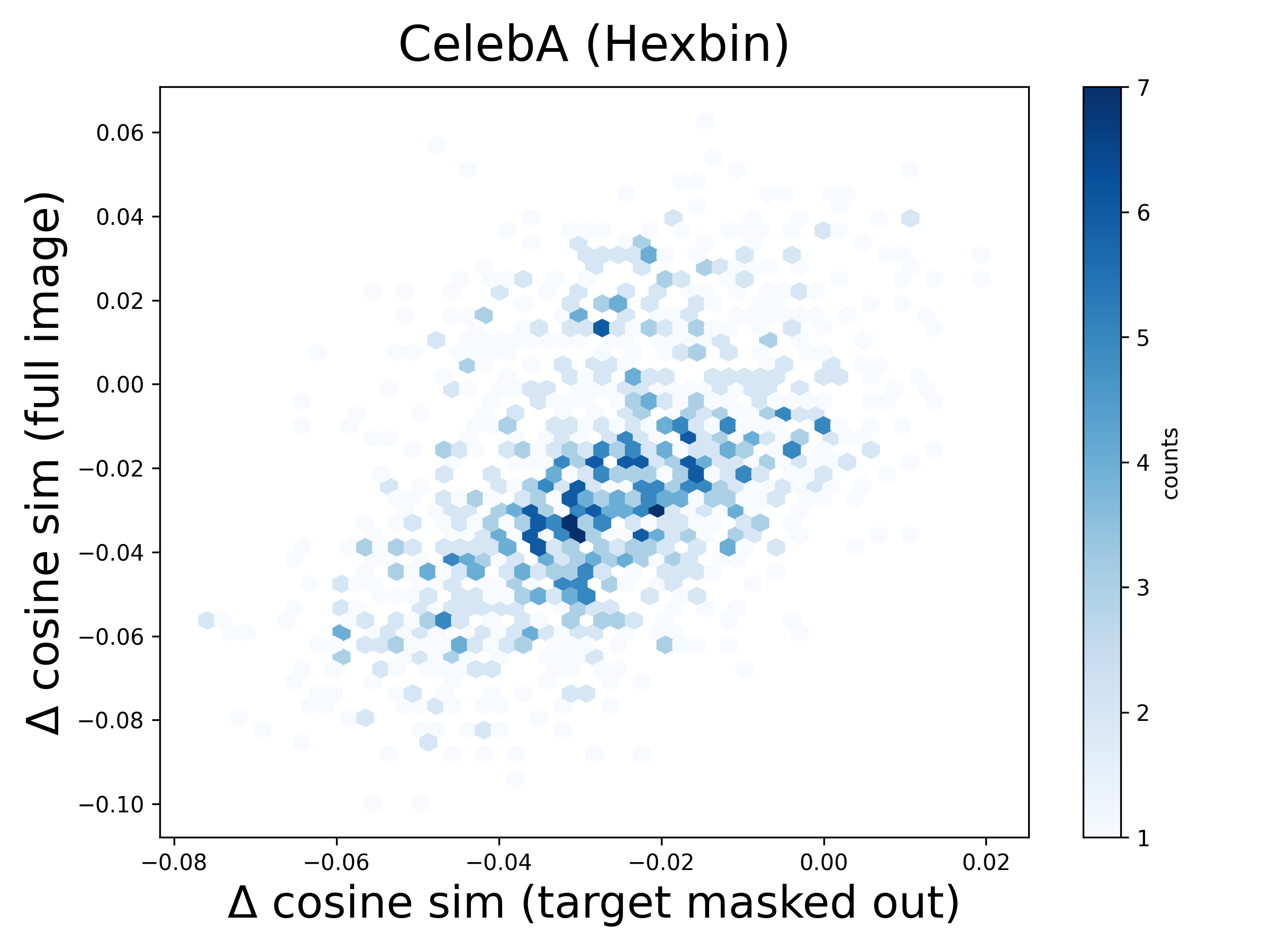} 
    \end{tabular}
  \end{center}
    \caption{Correlation between cosine similarity differences for non-target regions (x-axis: $\Delta_{\text{non-target}}$) and full images (y-axis: $\Delta_{\text{full}}$) with respect to two candidate text embeddings. Each point corresponds to an image randomly sampled from the dataset (1,500 total), illustrating how much the non-target region alone biases the prediction in the same direction as the full image across the Waterbirds~\cite{Sagawa20Waterbirds} and CelebA~\cite{CelebA} datasets.}
  \label{fig:correlation}
\end{figure}

Here, $E_{\text{prompt}_1}$ and $E_{\text{prompt}_2}$ correspond to the candidate text embeddings for each class. Plotting $\Delta{_\text{non-target}}$ against $\Delta_{\text{full}}$ across the dataset reveals a consistent positive correlation (as shown in Figure~\ref{fig:correlation}), suggesting that bias in non-target regions significantly influences the final prediction. Specifically, if the non-target region alone aligns more closely with a given class prompt, the full image embedding also tends to exhibit higher cosine similarity with the same class. These results highlight how contextually irrelevant regions can contaminate CLIP’s zero-shot predictions.

To mitigate the bias introduced by non-target regions, we propose a semantic region-guided strategy that isolates the target attribute (e.g., bird or hair), suppresses irrelevant regions, and recombines the features to produce more robust and semantically grounded predictions.

\subsection{SegDebias Framework}
SegDebias is a test-time, segmentation-guided debiasing pipeline that operates on any zero-shot classification task with an arbitrary set of text prompts \(\{\,\tau_i\}^{N}_{i=1}\).  Our approach proceeds in four steps:

\noindent
\textbf{1. Target Attribute Selection}
We define the \emph{target visual attribute} to be the semantic region most directly corresponding to the class labels (e.g.\ the \emph{bird} region in Waterbird vs.\ Landbird, or the \emph{hair} region for Blond Hair vs.\ Dark Hair).  By focusing on target regions, we avoid the combinatorial complexity of enumerating all potential confounding factors.
\noindent
\textbf{2. Promptable Segmentation.}  
To spatially isolate the target attribute identified in Step~1, we employ a two-stage segmentation process guided by vision-language models. First, we use a grounding detector such as Grounding DINO~\cite{liu2023grounding} with the prompt ``Photo of a \(\langle\)\textit{attribute}\(\rangle\)'' to generate a coarse bounding box around the relevant region. Then, within this region, we apply a promptable segmentation model to extract a fine-grained binary mask \(M_t\) that captures the target area. The complement of this mask, \(M_{\bar t}\), represents the non-target or background region and is used in subsequent debiasing steps.\\

\begin{figure}[H]
  \centering
  \includegraphics[width=\textwidth]{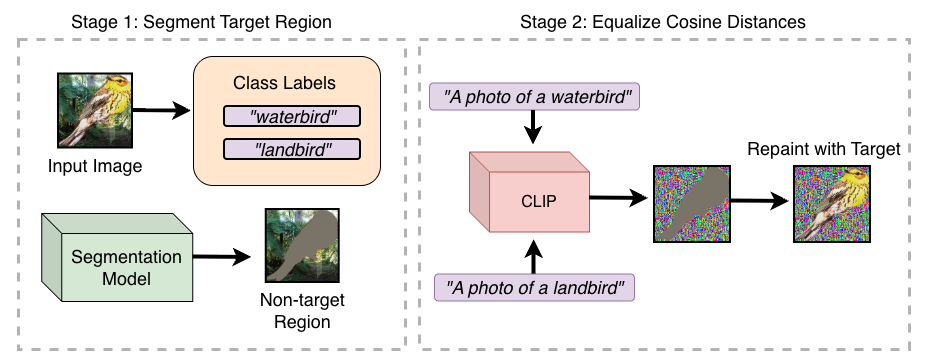}
  \vspace{0.5em} 
  \caption{Overview of the proposed debiasing pipeline for zero-shot image classification. Given an input image and associated candidate text embeddings, we identify the target attribute (e.g., bird), segment it from the image, and optimize the background (non-target attributes) to have equal cosine distances with all text embeddings. The target attribute is then repainted into the debiased background and passed to CLIP for final zero-shot prediction.}
  \label{fig:method}
\end{figure}

\noindent
\textbf{3. Non-Target Neutralization.}  
To reduce residual bias from the non-target region \(M_{\bar t}\), we retain the original image but constrain updates to this region only. Specifically, we compute the CLIP visual embedding from the partially masked image,
\(
  z_v \;=\; f_{\mathrm{CLIP}}(x'),
  \quad x' = x \odot M_{\bar t},
\)
and introduce a small perturbation \(\delta\) supported only on \(M_{\bar t}\) to decorrelate the non-target region from any specific class. Let \(\{z_{t_i}\}\) denote the text embeddings for each prompt \(\tau_i\). We then optimize \(\delta\) to equalize the cosine similarity between the perturbed visual embedding and all candidate text embeddings:
\begin{equation}
  \min_{\delta}\;\sum_{i<j}\left[\cos(z_v + \delta,\,z_{t_i}) - \cos(z_v + \delta,\,z_{t_j})\right]^{2}
  \quad\text{s.t.}\;\mathrm{supp}(\delta)\subseteq M_{\bar t}.
\end{equation}
This procedure ensures that the non-target visual attributes do not favor any particular class, effectively neutralizing spurious associations.\\

\noindent
\textbf{4. Reconstruction and Zero-Shot Classification.}  
Finally, we reconstruct the full image by restoring the original target region and blending it with the debiased background. Let \(x' = x \odot M_{\bar t}\) denote the background portion of the image and \(\delta\) the optimized perturbation supported on \(M_{\bar t}\). The final image is given by
\begin{equation}
\tilde{x} = (x \odot M_t) + (x' + \delta) \odot M_{\bar t}.
\label{eq:inpaint}
\end{equation}
We then compute the CLIP embedding \(f_{\mathrm{CLIP}}(\tilde{x})\) and perform zero-shot prediction via \(\arg\max_i\,\cos\big(f_{\mathrm{CLIP}}(\tilde{x}),\,z_{t_i}\big)\).

\section{Experiments}

\subsection{Datasets}

We evaluate \textbf{SegDebias} on two widely used datasets for studying spurious correlations in zero-shot classification: Waterbirds~\cite{Sagawa20Waterbirds} and CelebA~\cite{CelebA}. These benchmarks are commonly adopted in recent fairness and debiasing research~\cite{Dehdashtian24, zhang2022contrastive}, and we follow the same zero-shot evaluation setup as prior test-time debiasing work~\cite{Chuang23, Gerych24}.\\

\noindent
\textbf{Waterbirds}
\label{sec:4.1}
The Waterbirds dataset~\cite{Sagawa20Waterbirds} is a synthetic benchmark constructed from CUB-200-2011 by overlaying bird images onto either water or land backgrounds. The classification task is to distinguish between \emph{waterbird} and \emph{landbird}, but the background is spuriously correlated with the bird class, for example, most waterbirds appear over water. The smallest “worst-group” (waterbirds on land) comprises only 2\% of the training set, making the dataset a standard benchmark for evaluating worst-group robustness.\\

\noindent
\textbf{CelebA}  
CelebA~\cite{CelebA} is a large-scale facial attribute dataset with over 200{,}000 celebrity images annotated with 40 binary attributes. We focus on the binary hair color classification task, distinguishing \emph{blond hair} from \emph{non-blond hair}. Prior work~\cite{Serianni25, Chuang23, Dehdashtian24} has shown that CLIP often relies on gendered facial features (e.g., eyes or nose) when predicting hair color, leading to performance disparities across subgroups (e.g., blond females vs.\ males). We follow the standard train/validation/test splits used in FairerCLIP~\cite{Dehdashtian24} and DebiasedPrompt~\cite{Chuang23}.

\subsection{Implementation}

We evaluate \textbf{SegDebias} under the same test-time zero-shot protocol used by DebiasedPromp~\cite{Chuang23} and FairerCLIP~\cite{Dehdashtian24} on the Waterbirds~\cite{Sagawa20Waterbirds} and CelebA~\cite{CelebA} benchmarks. For Waterbirds, the target attribute is \emph{bird}, and we use the prompts \emph{“A photo of a waterbird”} and \emph{“A photo of a landbird”}. For CelebA, the target attribute is \emph{hair}, and we use \emph{“A photo of a celebrity with blond hair”} versus \emph{“…with dark hair”}. These minimal templates are fully compatible with zero-shot inference and focus the model on semantically relevant regions. All images are encoded using OpenAI's CLIP ViT-L/14. To verify model-agnostic effectiveness, we also validate on ViT-B/16 and ViT-B/32. For spatial localization, we use Grounding DINO~\cite{liu2023grounding} with the prompt “Photo of a \(\langle\)attribute\(\rangle\)” to obtain a coarse bounding box. To extract fine-grained masks, we use SAM-ViT-B~\cite{Kirillov23Segment} for Waterbirds, and a SegFormer-B5 face parsing model~\cite{Zhou21SegFormer} for CelebA.

We iteratively optimize a perturbation \(\delta\) over the non-target region \(M_{\bar t}\) to equalize cosine similarity across all text embeddings, stopping when the maximum pairwise difference falls below 0.001. All experiments are conducted in PyTorch on a single NVIDIA A100. Our perturbation-based debiasing procedure converges in under 70 ms on average. We report all performance metrics over three random seeds.

\subsection{Group Robustness}
We comparie SegDebias against approaches in both categories mentioned in Section \ref{sec:intro}.
\noindent
\textbf{Training-based methods (require labels or embedding fine-tuning):}
(1) ERM Linear~\cite{Kumar2022FineTuning}: a linear classifier trained with empirical risk minimization (ERM) on CLIP embeddings.  
(2) ERM Adapter~\cite{Gao2024CLIPAdapter}: a contrastive adapter trained via ERM.  
(3) WiSE-FT~\cite{Wortsman22WiseFT}: ensembling a fine-tuned ERM linear head with the original CLIP zero-shot classifier.  
(4) DFR (Up)~\cite{Kirichenko22DFR}: deep feature reweighting using upsampling.  
(5) CA~\cite{zhang2022contrastive}: a contrastive adapter trained on group-balanced data to improve robustness.  
(6) FairerCLIP~\cite{Dehdashtian24}: a projection-based method that debiases CLIP embeddings in RKHS to mitigate the influence of spurious relations.
\textbf{Test-time only methods (no additional training):}
(7) Zero-shot: standard CLIP zero-shot classification.  
(8) Orth-Proj~\cite{Chuang23}: removes spurious subspaces via orthogonal projection. 
(9) Orth-Cali~\cite{Chuang23}: orthogonal calibration with regularization to align prompt embeddings.
(10) BendVLM~\cite{Gerych24}: applies two-step, test-time debiasing by orthogonalizing to local attribute subspaces and equalizing similarity to reference images. We use gender and background as debiasing attributes for CelebA~\cite{CelebA} and Waterbirds~\cite{Sagawa20Waterbirds}, respectively.\\

\begin{table*}[h]
\centering
\small
\begin{tabular*}{.9\textwidth}{@{\extracolsep{\fill}} l|ccc|ccc }
\toprule
\multirow{2}{*}{Dataset} & \multicolumn{3}{c|}{Waterbirds} & \multicolumn{3}{c}{CelebA} \\
                        & \textbf{WG (\% $\uparrow$)} & \textbf{Avg (\% $\uparrow$)} & \textbf{Gap ($\downarrow$)}   & \textbf{WG (\% $\uparrow$)} & \textbf{Avg (\% $\uparrow$)} & \textbf{Gap ($\downarrow$)}   \\
\midrule
\multicolumn{7}{l}{\emph{Training‐based methods}}\\
ERM Adapter             & 78.4 & 97.8  & 19.4  & 36.7  & 94.2  & 57.5  \\
WiSE‐FT                 & 65.9 & 97.6  & 31.7  & 80.0  & 87.4  & 7.4   \\
DFR (Up)                & 65.9 & 96.1  & 30.2  & 83.7  & 91.2  & 7.5   \\
CA                      & 86.9 & 96.2  & 9.3   & 84.6  & 90.4  & 5.8   \\
FairerCLIP              & 78.1 & 85.1  & 7.1   & 86.1  & 88.0  & 1.9   \\
\midrule
\multicolumn{7}{l}{\emph{Test‐time only methods}}\\
Zero‐shot               & 45.3 & 84.4  & 39.1  & 72.8  & \textbf{87.6}  & 14.9  \\
Orth‐Proj.              & 61.4 & \underline{86.4}  & 25.0  & 71.1  & \underline{87.0}  & 15.9  \\
Orth‐Cali.              & \underline{68.8} & 84.5  & \textbf{15.7}  & \underline{76.1}  & 86.2  & \underline{10.1}\\
BendVLM  & 64.3 & 82.1  & 17.8  & 70.9  & 84.2  & 13.3  \\
SegDebias (Ours)        & \textbf{71.6} & \textbf{88.2} & \underline{16.6} & \textbf{81.6} & 85.1 & \textbf{3.5} \\
\bottomrule
\end{tabular*}
\vspace{1.2em}
\caption{Group robustness of VLMs with various debiasing methods using the ViT-L/14 backbone. Best and second-best results are shown in \textbf{bold} and \underline{underlined}, respectively.}
\label{tab:results}
\end{table*}

\noindent
\textbf{Evaluation Metrics.}  
We adopt metrics commonly used in group robustness studies, following prior work~\cite{Dehdashtian24}: Average Accuracy (Avg), Worst-Group Accuracy (WG), and the Gap. \textbf{Avg} reports the mean accuracy across all examples. \textbf{WG} denotes the precision of the worst performing group, typically the smallest or most underrepresented subgroup in the data set. \textbf{Gap} is defined as the difference between average and worst-group accuracy (Gap = Avg – WG), and quantifies disparity in performance across groups.\\

\noindent
\textbf{Empirical Results} As shown in Table \ref{tab:results}, SegDebias achieves the highest worst-group accuracy on both Waterbirds and CelebA, while also substantially narrowing the Avg–WG gap. In some cases, it even outperforms training-based methods, despite requiring no retraining or prior knowledge of spurious correlations. In CelebA, the overrepresentation of blond-haired female subjects in the test set (49\%) contributes to the slight decrease in average accuracy across all test-time debiasing methods, reflecting the removal of spurious shortcuts exploited by the zero-shot model.
\\
\begin{table}[h]
  \centering
  \small
  \begin{tabular}{lcccc}
    \toprule
    Backbone & Method            & \textbf{WG (\% $\uparrow$)} & \textbf{Avg (\% $\uparrow$)} & \textbf{Gap ($\downarrow$)} \\
    \midrule
    \multirow{2}{*}{ViT‐B/16} 
      & Zero‐shot         & 25.8     & \textbf{79.8}      & 54.0      \\
      & SegDebias  & \textbf{36.4}     & 74.8      & \textbf{38.4}      \\
    \midrule
    \multirow{2}{*}{ViT‐B/32} 
      & Zero‐shot         & 41.5     & \textbf{69.1}      & 27.6      \\
      & SegDebias & \textbf{50.2}     & 61.1      & \textbf{10.9}      \\
    \bottomrule
  \end{tabular}
  \vspace{1.1em}
\caption{Group robustness metrics of \textsc{SegDebias} vs.\ zero-shot CLIP on the Waterbirds dataset using ViT-B/16 and ViT-B/32 backbones.}
  \label{tab:backbone-ablation}
\end{table}

\noindent
\textbf{ViT Backbones} To assess the robustness of SegDebias across different patch resolutions, we evaluated two CLIP backbones, ViT-B/16 and ViT-B/32, on the Waterbirds dataset (Table~\ref{tab:backbone-ablation}). In both settings, SegDebias consistently improves worst-group accuracy and substantially narrows the accuracy gap, highlighting its effectiveness across backbone variants.

\subsection{Attention‐IoU Evaluation}

\noindent
\textbf{Attention Map Extraction and Evaluation.}  
To assess the alignment between CLIP’s ViT encoder attention and semantically relevant regions, we extract attention maps and evaluate them against ground-truth masks using the Attention-IoU metric by Serianni et al.~\cite{Serianni25}.

In ViT\cite{Dosovitskiy21ViT}, the CLS token summarizes global context, and its attention reveals which regions influence the model’s prediction. We extract an attention map \(\hat A \in [0,1]^{G \times G}\) by collecting the attention weights from the CLS token to all patch tokens at each transformer block. These weights are averaged across all heads and layers, then reshaped into a spatial \(G \times G\) grid and normalized to unit range. The resulting map \(\hat A\) highlights regions most influential to the final CLS embedding. Given a ground-truth binary mask \(M_t\) for the target attribute, we define the predicted attention region as the smallest set of patches that cumulatively capture 50\% of the total attention mass. The Attention-IoU score is then computed as:

\begin{equation}
  \mathrm{Attention\text{-}IoU} \;=\;
  \frac{\lvert\,\{p:\hat A(p)\ge\tau\}\,\cap\,M_t\rvert}
       {\lvert\,\{p:\hat A(p)\ge\tau\}\,\cup\,M_t\rvert}\!,
  \quad
  \tau = \min\Bigl\{\tau':\sum_{p:\hat A(p)\ge\tau'}\hat A(p)\ge 0.5\Bigr\}.
  \label{eq:attention-iou}
\end{equation}

This single-threshold formulation yields a concise measure of spatial overlap between the model's high-attention regions and the true semantic region of interest.

\begin{figure*}[h]
  \centering
  \begin{tabular}{@{}c@{\hspace{4pt}}c@{\hspace{12pt}}c@{\hspace{4pt}}c@{\hspace{12pt}}c@{\hspace{4pt}}c@{}}
    % Waterbird pairs
    \includegraphics[width=0.14\textwidth]{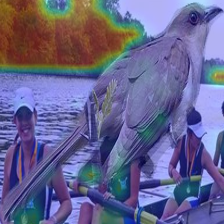} &
    \includegraphics[width=0.14\textwidth]{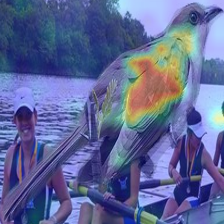} &
    \includegraphics[width=0.14\textwidth]{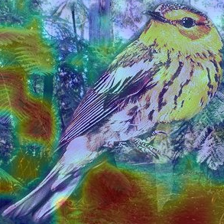} &
    \includegraphics[width=0.14\textwidth]{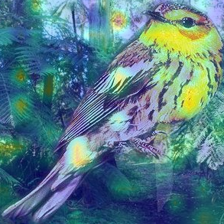} &
    \includegraphics[width=0.14\textwidth]{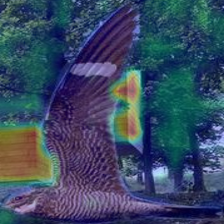} &
    \includegraphics[width=0.14\textwidth]{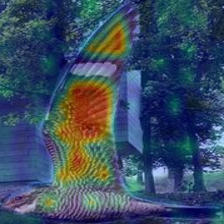} \\
    % CelebA pairs
    \includegraphics[width=0.14\textwidth]{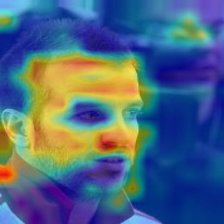} &
    \includegraphics[width=0.14\textwidth]{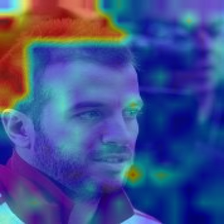} &
    \includegraphics[width=0.14\textwidth]{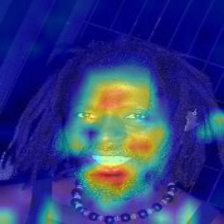} &
    \includegraphics[width=0.14\textwidth]{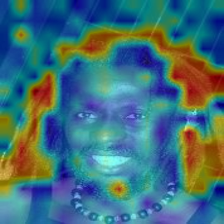} &
    \includegraphics[width=0.14\textwidth]{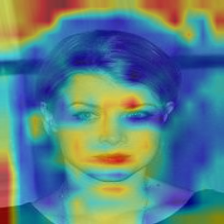} &
    \includegraphics[width=0.14\textwidth]{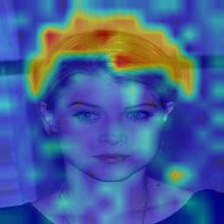}
  \end{tabular}
  \vspace{1em}
  \caption{Comparison of attention maps before (left) and after (right) applying SegDebias on two tasks. Top row: three pairs of waterbird examples; bottom row: three pairs of CelebA hair examples. In each pair, vanilla CLIP’s attention (left) tends to scatter over confounding background or depiction biases, whereas SegDebias’s attention (right) aligns more closely with the true semantic region, yielding higher interpretable focus.}
  \label{fig:before-after-attn}
\end{figure*}

\noindent
\begin{minipage}[c]{0.58\columnwidth}
\textbf{Results} Table~\ref{tab:attn-iou} reports Attention‐IoU on Waterbirds and CelebA for both zero‐shot CLIP and SegDebias.  By neutralizing non-target features before recombining with the target region, SegDebias refocuses the model’s attention onto the semantic attribute of interest, yielding a marked increase in Attention‐IoU and demonstrating more faithful and interpretable behavior as shown in Figure \ref{fig:before-after-attn}.
\end{minipage}%
\hfill
\begin{minipage}[c]{0.38\columnwidth}
  \centering\small
  \setlength{\tabcolsep}{2pt}  % tighten columns
  \renewcommand{\arraystretch}{0.9}  
  \begin{tabular}{@{}lcc@{}}
    \toprule
     & Waterbirds & CelebA \\
    \midrule
    Zero‐shot   & 70.0\% & 52.9\% \\
    SegDebias   & \textbf{79.9\%} & \textbf{66.8\%} \\
    \bottomrule
  \end{tabular}
  \vspace{1em}
  \captionof{table}{Attention-IoU mask score with Clip ViT-L14.}
  \label{tab:attn-iou}
\end{minipage}

\subsection{Ablation Studies}
\begin{table}[h]
  \centering
  \small
  \setlength{\tabcolsep}{6pt}
  \begin{tabular}{@{}lcccc@{}}
    \toprule
    Metric         & Target-Only & Noise-Filled & Random Repaint & \textbf{SegDebias} \\
    \midrule
    WG (\% $\uparrow$)        & 56.5        & 33.9         & 64.3           & \textbf{71.6} \\
    Avg (\% $\uparrow$)       & \textbf{88.9}        & 60.2         & 86.2           & 88.2 \\
    Gap (\% $\downarrow$)       & 32.4        & 18.3         & 21.9           & \textbf{16.6} \\
    \bottomrule
  \end{tabular}
  \vspace{1em}
  \caption{Ablation on different variant inputs, experimented with Waterbirds~\cite{Sagawa20Waterbirds} dataset.}
  \label{tab:ablation-wb}
\end{table}
\vspace{-0.2em}
\noindent
To demonstrate the effect of non-target attribute debiasing, we compare four input variants on the Waterbirds dataset in a single pass:  
(1) \textbf{Target‐Only Input} retains only the segmented bird region, masking all background;  
(2) \textbf{Noise‐Filled Background} restores bird pixels but fills the non‐target region with Gaussian noise (\(\theta\)=0.1) without further optimization;  
(3) \textbf{Random Repaint} repaints the target region onto an unoptimized, randomly‐sampled noisy background; and  
(4) \textbf{Full SegDebias}. As shown in Table~\ref{tab:ablation-wb}, while both noise-filled and random-repaint variants reduce spurious cues, the unoptimized background still introduces residual bias that can favor one class. In contrast, SegDebias explicitly balances background similarity, leading to the best group fairness metrics in both WG and Gap.

\section{Discussion}

\noindent
\textbf{Effectiveness on Transformer Based Backbones.}  
SegDebias demonstrates strong empirical gains when applied to ViT based CLIP models, aligning with the increasing use of transformers in vision and language representation learning~\cite{Dosovitskiy21ViT, zhai2021scaling}. Unlike convolutional networks, which operate through local receptive fields, transformer models tokenize the image into patches and process them using global self attention. This structure enables ViTs to integrate spatially distant information and attend to object level context across the entire image, making them well suited for masking or region specific interventions. SegDebias leverages this property by neutralizing non target patches while preserving discriminative attributes, which improved group robustness. Our results further support observations that ViTs exhibit a form of semantic grounding through attention that can be modulated at test time~\cite{raghu2021vision, chefer2021transformer}. Because each patch embedding contributes globally to the final representation, masking a region does not necessarily break feature continuity. This contrasts with convolutional pipelines, where localized masking can fragment spatial dependencies. As ViTs~\cite{Dosovitskiy21ViT} continue to serve as the foundation for modern vision language systems, segmentation guided strategies like ours offer a promising path for addressing fairness and robustness in a model agnostic and non intrusive way.

\section{Conclusion}
We propose SegDebias, a segmentation-guided test-time debiasing method for ViT-based CLIP models that requires no representation tuning or prior knowledge of the test distribution. By isolating target attributes and suppressing spurious regions, SegDebias enhances group robustness and attention alignment over existing test-time debiasing methods. Future work could extend SegDebias to abstract attributes and complex biases like multi-attribute entanglement. Integrating self-supervised localization may further enable the model to dynamically verify whether attended regions are semantically relevant during test time.

\bibliography{egbib}
\end{document}